\documentclass[journal]{IEEEtran}
\usepackage[utf8]{inputenc}
\usepackage[T1]{fontenc}
\usepackage{graphicx}
\usepackage{cite}
\usepackage{amsmath,amssymb,amsfonts}
\usepackage{algorithm}
\usepackage{algorithmic}
\usepackage{booktabs}
\usepackage{multirow}
\usepackage{siunitx}
\usepackage{url}
\usepackage{balance}
\usepackage{flushend}
\usepackage{textcomp}
\usepackage{xcolor}
\usepackage{tikz}
\usepackage{ragged2e}
\usepackage{microtype}
\usepackage{acronym}
\usepackage{tcolorbox}
\usepackage{hyperref}
\hypersetup{
    hidelinks,
    colorlinks=false,
    linkcolor=black,
    citecolor=black,
    urlcolor=black
}

\title{Nauplius Optimisation for Autonomous Hydrodynamics}

\author{\IEEEauthorblockN{Shyalan Ramesh, Scott Mann, Alex Stumpf\\}
\IEEEauthorblockA{\textit{School of Computing, Engineering, and Mathematical Sciences} \\
\textit{La Trobe University} \\
Melbourne, Australia \\
\{Shyalan.Ramesh, S.Mann, A.Stumpf\}@latrobe.edu.au}
}

\begin{document}

\maketitle

\begin{abstract}
Autonomous Underwater vehicles must operate in strong currents, limited acoustic bandwidth, and persistent sensing requirements where conventional swarm optimisation methods are unreliable. This paper formulates an irreversible hydrodynamic deployment problem for \ac{AUV} swarms and presents \ac{NOAH}, a novel nature-inspired swarm optimisation algorithm that combines current-aware drift, irreversible settlement in persistent sensing nodes, and colony-based communication. Drawing inspiration from the behaviour of barnacle nauplii, \ac{NOAH} addresses the critical limitations of existing swarm algorithms by providing hydrodynamic awareness, irreversible anchoring mechanisms, and colony-based communication capabilities essential for underwater exploration missions. The algorithm establishes a comprehensive foundation for scalable and energy-efficient underwater swarm robotics with validated performance analysis. Validation studies demonstrate an 86\% success rate for permanent anchoring scenarios, providing a unified formulation for hydrodynamic constraints and irreversible settlement behaviours with an empirical study under flow.
\end{abstract}

\begin{IEEEkeywords}
Swarm optimisation, underwater robotics, bio-inspired algorithms, autonomous underwater vehicles, hydrodynamic adaptation, irreversible settlement, colony communication, \ac{NOAH} algorithm, marine robotics.
\end{IEEEkeywords}

\section{Introduction}
Underwater exploration and environmental monitoring are increasingly crucial in marine conservation, offshore resource management and scientific research. The aquatic domain presents formidable challenges due to complex hydrodynamics, limited communication and severe energy limitations \cite{Hasan2024}. 

Swarm robotics with \ac{AUV}s offers improved coverage, fault tolerance, and adaptability for such tasks \cite{connor2021current}. Multiple \ac{AUV}s operating collectively can survey larger areas faster and withstand single-node failures better than a single vehicle \cite{cai2024multi}. Recent advances in multi-AUV exploration underscore these benefits, with new distributed strategies achieving robust, wide-area ocean surveys \cite{mu2025coverage}. 

Wang \textit{et al.} demonstrate autonomous cooperative systems for underwater target detection and localisation, highlighting the potential for coordinated multi-\ac{AUV} operations \cite{wang2023multi}. Wen \textit{et al.} address intelligent decision-making for the planning of the \ac{AUV} path against ocean current disturbances \cite{wen2024intelligent}.

However, swarm optimisation algorithms, originally designed for terrestrial or aerial robots, often fail in aquatic settings due to three critical limitations: difficulty modelling dynamic ocean currents, lack of sustained sensing mechanisms for long-term observation, and challenges with acoustic communication bandwidth and latency constraints \cite{connor2021current}. Standard optimisers such as \ac{PSO} and \ac{ACO} assume fully mobile agents and simple communication \cite{Lhotska2006}, neglecting the fluid forces and intermittent acoustic links that \ac{AUV} swarms face \cite{loncar2019heterogeneous}. Moreover, existing multi-AUV systems rarely allow vehicles to become stationary observers, with irreversible anchoring largely absent except in heterogeneous specialised networks \cite{loncar2019heterogeneous}. Although emerging research addresses individual aspects such as energy-efficient \ac{AUV} deployment \cite{xu2025energy} and ocean flow-based coverage planning \cite{mu2025coverage}, no previous work has merged hydrodynamic drift adaptation, persistent anchoring, and colony-style communication into a unified framework. In this context, we address irreversible hydrodynamic deployment by converting mobile \ac{AUV} swarms into permanently anchored sensing and relay colonies under ocean flow, and we validate this new \ac{NOAH} framework through anchoring accuracy, convergence, and comparative performance experiments.

\section{Related Work}

\begin{figure}[htbp]
    \centering
    \includegraphics[width=\columnwidth]{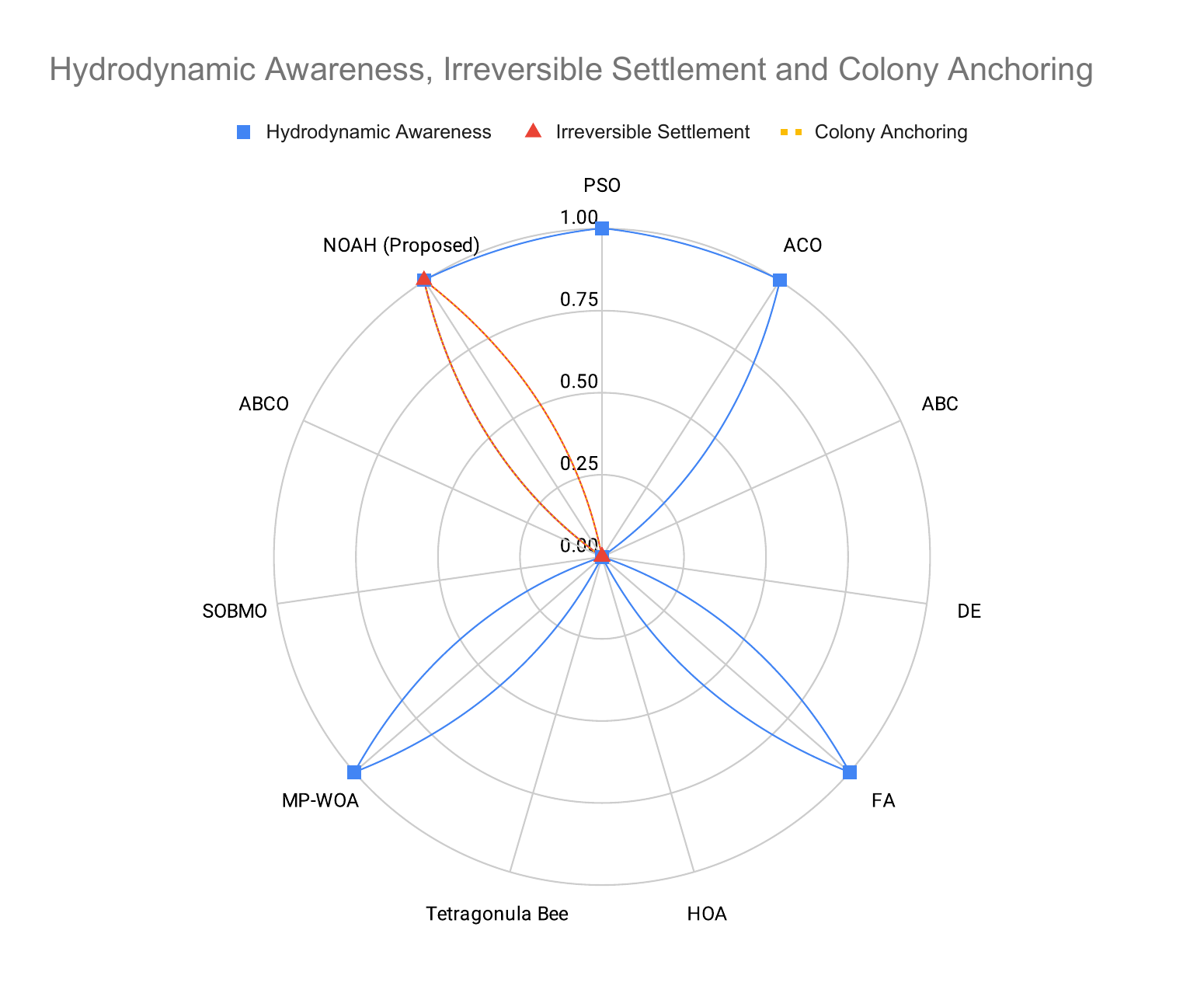}
    \caption{Research gaps in current swarm optimisation capabilities for underwater exploration and marine robotics applications.}
    \label{fig:research_gaps}
\end{figure}

Swarm intelligence has led to a wide family of metaheuristic optimisation methods that have been successfully applied across domains, including underwater robotics. These approaches generally fall into two categories: foundational algorithms, which established the principles of swarm-based optimisation, and recent bio-inspired algorithms, which introduce new behavioural dynamics. This section reviews both groups, highlighting their contributions and applications in aquatic missions. Table~\ref{tab:comparison_table} provides a detailed comparison of these algorithms, while Fig.~\ref{fig:research_gaps} identifies the critical capability gaps that existing approaches do not address.

\begin{table}[htbp]
\centering
\caption{Comparison of swarm optimisation algorithms for robotics applications.}
\label{tab:comparison_table}
\resizebox{\columnwidth}{!}{%
\begin{tabular}{@{}>{\raggedright\arraybackslash}p{2.5cm}>{\raggedright\arraybackslash}p{2.3cm}>{\raggedright\arraybackslash}p{3.9cm}>{\raggedright\arraybackslash}p{1.1cm}@{}}
\toprule
\textbf{Algorithm} & \textbf{Inspiration} & \textbf{Mechanism} & \textbf{Reference} \\
\midrule
\ac{PSO} & Flocking (birds/fish) & Velocity + best memory & \cite{kennedy1995particle}, \cite{panda2020comprehensive} \\
\ac{ACO} & Ant foraging & Pheromone-guided path search & \cite{dorigo2006ant}, \cite{mai2023uav} \\
\ac{ABC} & Bee foraging & Employed/onlooker/scout phases & \cite{karaboga2005idea}, \cite{contreras2015mobile} \\
\ac{DE} & Evolutionary vectors & Mutation \& recombination & \cite{storn1997differential} \\
\ac{FA} & Firefly attraction & Distance-based movement & \cite{yang2010firefly} \\
\ac{HOA} & Hippopotamus behaviour & Trinary-phase search & \cite{amiri2024hippopotamus} \\
Tetragonula Bee & Hive construction & Distributed coordination & \cite{gamez2025novel} \\
\ac{MP-WOA} & Whale predation & Encircling \& spiral & \cite{chakraborty2021novel} \\
\ac{SOBMO} & Barnacle mating & Opposition-based search & \cite{ahmed2024selective} \\
\ac{ABCO} & Bacterial foraging & Explore-exploit-reproduce & \cite{kogam2025adaptive} \\
\ac{NOAH} (Proposed) & Barnacle larvae on whales & Drift + irreversible settlement + colony communication & This work \\
\bottomrule
\end{tabular}%
}\end{table}
\vspace{-0.5em}

\subsection{Foundational Swarm Optimisation Methods}

Foundational swarm algorithms have established the core principles of collective intelligence, but exhibit notable limitations for aquatic environments. Particle Swarm Optimisation (\ac{PSO})~\cite{kennedy1995particle} simulates flocking behaviour by updating positions based on the best individual and global positions, finding applications in \ac{AUV} path planning~\cite{panda2020comprehensive}. Recent work by Li \textit{et al.} demonstrates improved \ac{PSO} for \ac{AUV} 3D path planning \cite{li2025auv}, while Sun \textit{et al.} address energy-optimised path planning with ocean current considerations \cite{sun2022energy}. Ant Colony Optimisation (\ac{ACO})~\cite{dorigo2006ant} uses pheromone-mediated foraging for efficient energy routing. Recent advances in dual-strategy \ac{ACO} algorithms, originally developed for \ac{UAV} path planning~\cite{mai2023uav}, demonstrate improved convergence and 3D navigation capabilities that can be adapted for \ac{AUV} applications in underwater environments, with Yuan \textit{et al.} integrating \ac{ACO} with dynamic window algorithms for communication efficiency \cite{yuan2025integrated}. Ronghua \textit{et al.} present an improved \ac{ACO} for the safe planning of \ac{AUV} paths \cite{ronghua2024improved}. Artificial Bee Colony (\ac{ABC})~\cite{karaboga2005idea} divides agents into employed, onlooker, and scout bees for the \ac{AUV} way point traversal~\cite{contreras2015mobile}, with Prasath \textit{et al.} applying Enhanced \ac{ABC} algorithms for underwater image enhancement \cite{Prasath2020}. Differential Evolution (\ac{DE})~\cite{storn1997differential} uses mutation and recombination for \ac{AUV} rendezvous planning~\cite{panda2020comprehensive}, while Chen \textit{et al.} address finite-time velocity-free rendezvous control \cite{chen2022finite}. The Firefly Algorithm (\ac{FA})~\cite{yang2010firefly} models the attraction based on brightness to plan underwater paths~\cite{panda2020comprehensive}.

Despite their effectiveness in terrestrial applications, these foundational swarm intelligence methods have been adapted for underwater exploration \cite{Aghababa2012}, but lack essential marine-specific capabilities: hydrodynamic flow field adaptation, irreversible settlement mechanisms, colony-based communication protocols and energy-efficient anchoring strategies.

\subsection{Contemporary Bio-Inspired Approaches}

Recent biologically-motivated algorithms introduce novel behavioural metaphors but share fundamental limitations with foundational methods. The Hippopotamus Optimisation Algorithm (\ac{HOA})~\cite{amiri2024hippopotamus} models the positioning, defence, and evasion phases, while the Tetragonula Bee Hive Optimisation~\cite{gamez2025novel} emphasises distributed coordination through hive construction behaviours. Enhanced whale optimisation~\cite{chakraborty2021novel} strengthens encircling and spiral search patterns, with Yan \textit{et al.} applying whale optimisation for \ac{AUV} path planning \cite{yan2021two} and Nadimi-Shahraki \textit{et al.} providing systematic reviews of whale optimisation algorithms \cite{nadimi2023systematic}. Selective opposition-based barnacle mating optimisation (\ac{SOBMO})~\cite{ahmed2024selective} employs barnacle-inspired mating strategies, while Darvishpoor \textit{et al.} review nature-inspired algorithms from oceans to space \cite{darvishpoor2023nature}. Adaptive bacterial colony optimisation (\ac{ABCO})~\cite{kogam2025adaptive} combines exploration, exploitation, and reproduction phases for accelerated convergence, and Kaveripakam \textit{et al.} applied clustering-based dragonfly optimisation for underwater wireless sensor networks \cite{kaveripakam2023clustering}.

These contemporary methods share critical limitations: agents remain fully mobile throughout optimisation, lack hydrodynamic modelling capabilities, and do not provide irreversible settlement mechanisms. Although \ac{SOBMO}'s barnacle inspiration represents the closest biological connection to \ac{NOAH}'s approach, it focuses on mating behaviour rather than settlement and environmental adaptation.

\subsection{Recent Advances in Underwater Swarm Robotics}

Recent developments in underwater swarm robotics have focused on addressing marine environment challenges, although gaps remain in persistent monitoring and environmental adaptation. Connor et al.~\cite{connor2021current} provide a comprehensive review highlighting critical limitations in acoustic communication bandwidth and latency constraints. Cai et al.~\cite{cai2024multi} introduce multi-\ac{AUV} distributed collaborative search methods, while Lončar et al.~\cite{loncar2019heterogeneous} present heterogeneous robotic swarms incorporating static nodes alongside mobile \ac{AUV}s. Xu et al.~\cite{xu2025energy} address the energy-efficient 3D deployment of \ac{AUV}-enabled mobile relays, and Mu and Gao~\cite{mu2025coverage} develop multi-\ac{AUV} coverage planners that account for ocean flow and sonar constraints.

However, these approaches share common limitations: they focus on mobile coordination without persistent monitoring capabilities, rely on specialised static infrastructure that limits deployment flexibility, or remain fully mobile throughout missions without irreversible settlement mechanisms. Collectively, they do not address the fundamental need for persistent monitoring and environmental adaptation that \ac{NOAH} provides through its unified framework.

\subsection{Comparative Analysis}

\begin{figure}[htbp]
    \centering
    \includegraphics[width=\columnwidth]{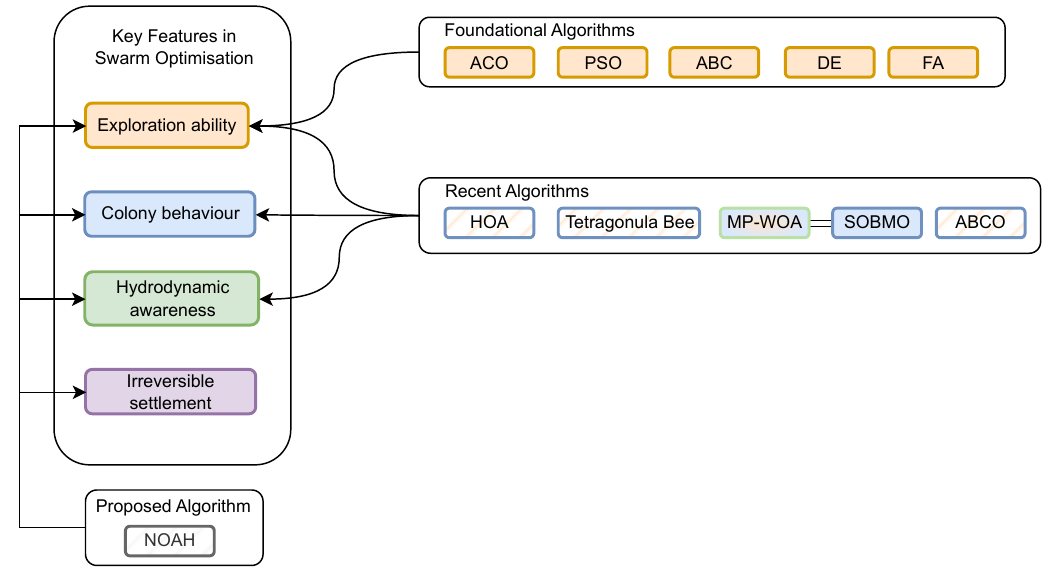}
    \caption{Feature comparison of swarm optimisation algorithms highlighting unique capabilities and limitations.}
    \label{fig:comparison_diagram}
\end{figure}

Table~\ref{tab:comparison_table} summarises the contrast between foundational and contemporary swarm algorithms, while Fig.~\ref{fig:comparison_diagram} categorises algorithms according to four critical features: exploration capacity, colony behaviour, hydrodynamic awareness, and irreversible settlement. The central intersection representing all four features is empty, demonstrating that no existing algorithm satisfies these combined requirements.

The irreversible settlement mechanism represents a fundamental paradigm shift from conventional swarm algorithms. Unlike traditional approaches in which agents remain fully mobile throughout missions, irreversible settlement enables agents to transition from exploration to exploitation by permanently anchoring at optimal locations. This capability is particularly crucial for underwater applications, where persistent monitoring nodes are essential for long-term environmental observation, data collection, and communication relay functions. The irreversible nature ensures that discovered optimal locations are preserved and cannot be lost due to subsequent exploration, creating a progressive freeze strategy that guarantees convergence while maintaining the discovered optima.

As demonstrated in Fig.~\ref{fig:research_gaps} and Table~\ref{tab:comparison_table}, \ac{NOAH} provides a unified framework that integrates hydrodynamic drift adaptation, irreversible settlement, and colony-mediated communication to address identified limitations in underwater swarm robotics.

\section{Background}

To address these limitations, we propose \textbf{\ac{NOAH} (Nauplius Optimisation for Autonomous Hydrodynamics)}, a novel nature-inspired swarm algorithm designed specifically for marine robotics applications. The algorithm draws inspiration from barnacle nauplii that drift with currents, settle irrevocably in optimal hosts, and form communicating colonies \cite{nogata2006larval}. By fusing these biologically-motivated behaviours, \ac{NOAH} enables a swarm of \ac{AUV}s to dynamically adapt to ocean currents, anchor certain members as persistent monitoring nodes, and maintain a robust networked colony.

\begin{figure}[htbp]
    \centering
    \includegraphics[width=0.7\columnwidth]{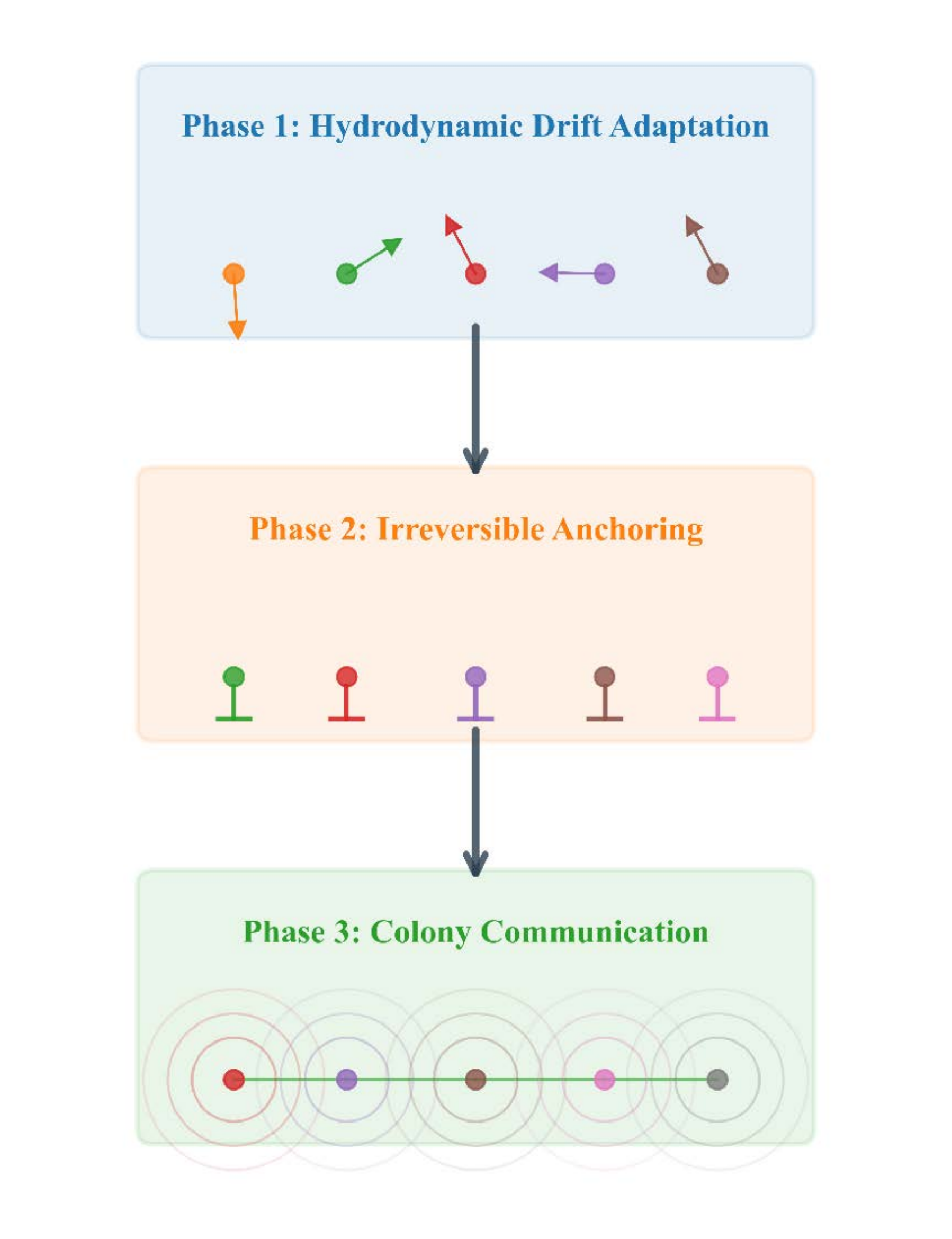}
    \caption{\ac{NOAH} three-phase methodology framework: Hydrodynamic Drift Adaptation, Irreversible Anchoring, and Colony Communication.}
    \label{fig:noah_methodology}
\end{figure}

The methodology operates through three distinct phases, as illustrated in Fig.~\ref{fig:noah_methodology}. This paper contributes a novel algorithm that unifies these capabilities in a \ac{AUV} swarm, along with a detailed mathematical foundation that addresses critical limitations in existing swarm algorithms for underwater applications.

\begin{tcolorbox}[colback=blue!5!white,colframe=blue!50!black,title=Problem Formulation]
Unlike classical \ac{PSO}/\ac{ACO} which aim to continuously move agents toward optima, our goal is to solve an irreversible hydrodynamic deployment problem by converting mobile swarm agents into stationary sensor nodes under hydrodynamic drift constraints. This transforms the problem from continuous mobility optimisation to irreversible deployment under ocean currents.
\end{tcolorbox}

\section{Biological Inspiration: Barnacle Nauplii on Whales}

Marine environments present fundamentally different challenges compared to terrestrial systems, requiring algorithms that can exploit hydrodynamic forces rather than resist them. Barnacle nauplii have evolved sophisticated strategies for navigating these challenges through irreversible commitment mechanisms that ensure survival in dynamic ocean currents.

\subsection{Understanding Nauplius Behaviour}
Barnacle nauplii undergo a critical developmental transition from free-swimming larvae to cyprid larvae that exhibit irreversible settlement behaviour \cite{nogata2006larval}. This irreversible anchoring mechanism is essential for survival, as cyprids secrete permanent cement that permanently bonds them to suitable substrates \cite{Kotsiri2018}. The irreversible nature of settlement ensures that once a cyprid commits to a location, it cannot relocate, making settlement decisions crucial for long-term survival and reproductive success. This biological constraint provides fundamental insights for designing underwater robotic systems that must balance exploration with permanent deployment strategies in dynamic marine environments where relocation is energetically costly and communication is intermittent \cite{wright1999influence}.

\subsection{Larval Drift and Hydrodynamic Cues}
Barnacle larvae progress through six naupliar stages before reaching the cyprid phase, consistent with observations in both intertidal and epizoic barnacle species. The development time is temperature-dependent, with cyprids emerging in approximately 7 to 8 days at 20 to 25 °C, followed by settlement and metamorphosis reaching juvenile form on day 12 to 18. During this phase of planktonic drift, larvae are passively transported by ambient currents \cite{nogata2006larval}, enabling broad dispersal before settlement.

When approaching large marine hosts, hydrodynamics strongly shapes micro-habitats of encounter and settlement. The larvae actively swim to exploit the favourable flow structures for settlement \cite{larsson2016instantaneous}. Recent work by Reidenbach \textit{et al.} demonstrates hydrodynamic interactions with coral topography and their impact on larval settlement \cite{reidenbach2021hydrodynamic}, while Ella and Genin examine capture dynamics at different flow speeds \cite{ella2024capture}. 

\begin{figure}[htbp]
    \centering
    \includegraphics[width=1\linewidth]{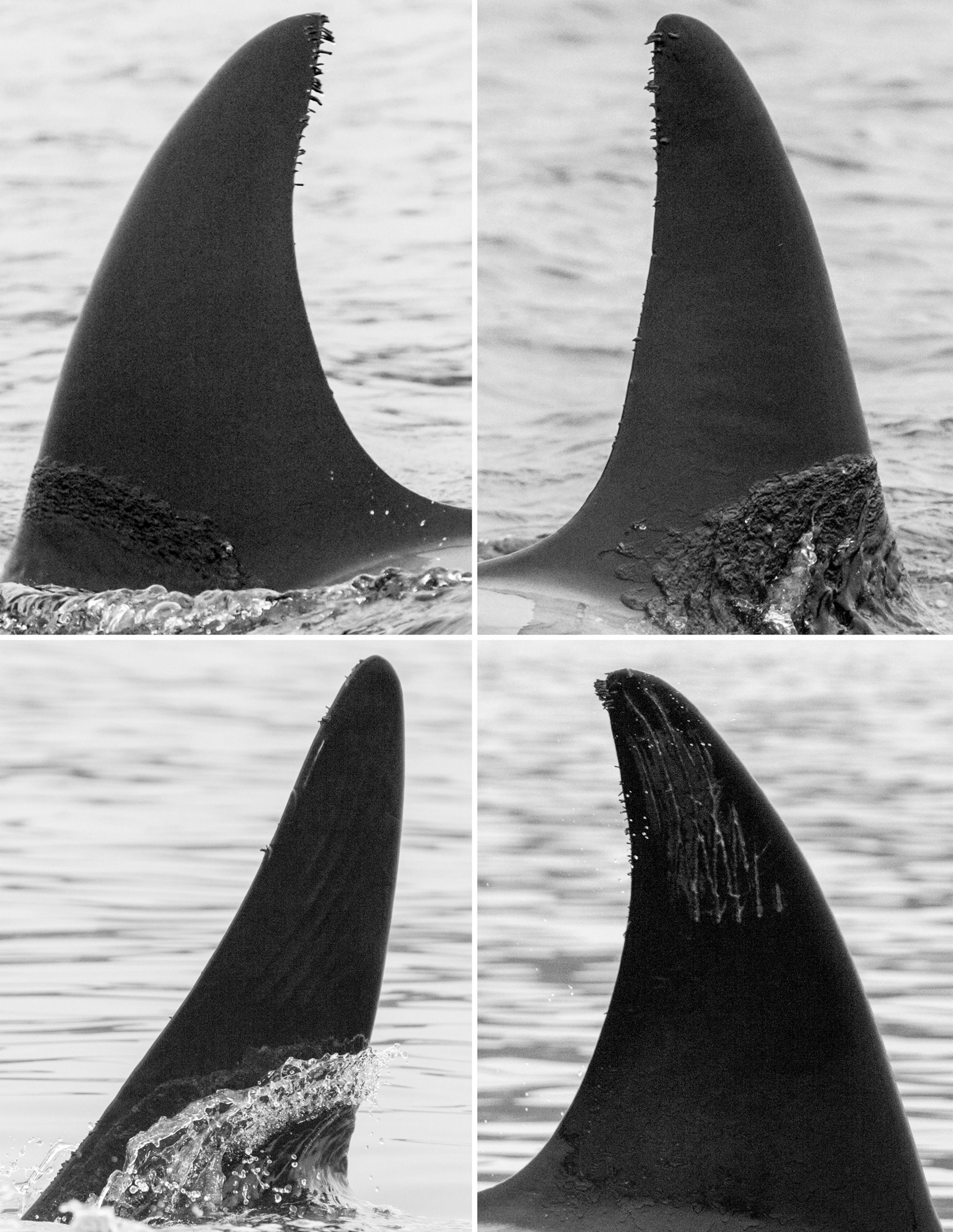}
    \caption{Photographs of four killer whales (O. orca) observed off northern Baffin Island in the eastern Canadian Arctic with Xenobalanus barnacles attached to the trailing edge of their upper dorsal fins. Adapted from Matthews \cite{matthews2020epizoic}.}
    \label{fig:whalebarnacles}
\end{figure}

Analyses on small cetaceans show that the commensal barnacle \textit{Xenobalanus globicipitis} concentrates along the trailing edges of the appendages, with preference for the centre of the fluke. In delphinids, the densities of the dorsal side exceed the densities of the ventral side, consistent with asymmetric fluke kinematics and adhesion-related shear forces \cite{carrillo2015living, Pugliese2012}. The trailing-edge colonisation pattern can be observed directly on the dorsal fins of killer whales, as shown in Fig.~\ref{fig:whalebarnacles}.

\subsection{Host Selection and Settlement}
Beyond hydrodynamic guidance, chemical confirmation is a key determinant of settlement. Recent work identified a waterborne settlement pheromone for barnacles as \textit{adenosine}. Settlement activity remains robust to experimental warming and acidification, underscoring the reliability of chemical signalling under fluctuating conditions \cite{Wu2024}. 

Recent studies demonstrate that exogenous adenosine significantly increases cyprid settlement rates and exploration frequency. It also improves the mechanical properties of adhesive footprints \cite{xu2025exogenous}. Liu \textit{et al.} examine chemical signalling in biofilm-mediated bio-fouling \cite{liu2024chemical}, providing information on settlement mechanisms. 

In practice, cyprids integrate hydrodynamic concentration near suitable micro-habitats with host-derived chemical signals \cite{wright1999influence}. Settlement occurs only when cue strength and contact time exceed a threshold, as anchoring is irreversible.

\subsection{Colony Formation and Influence}
Following anchoring, local hydrodynamics and conspecific signalling reinforce gregarious settlement. This drives colony growth along flow-favourable edges \cite{mignucci2022barnacles}. Positive feedback improves reproductive opportunity and feeding efficiency through consistent flow delivery. It imposes negligible drag costs for large hosts compared to body mass. 

Host specificity and adaptive evolution in settlement behaviour further enhance colony formation, as demonstrated in coral-associated barnacle larvae \cite{yap2023host}. Colonisation patterns such as those shown in Fig.~\ref{fig:whalebarnacles} provide the biological basis for \ac{NOAH}'s anchoring and colony-communication modules.

\section{The NOAH Algorithm}

We present the mathematical foundation for \ac{NOAH}, integrating the three core capabilities outlined in Section II.

\subsection{State and Fields}
\ac{NOAH} models each agent \ac{AUV} after a barnacle larva (nauplius/cyprid) drifting and seeking a settlement site. Each agent $i$ at time $t$ is characterised by three key state variables. The agent's position $x_{i,t} \in [0,1]^d$ represents its location in the $d$-dimensional search space, normalised to a unit hypercube. The velocity $v_{i,t} \in \mathbb{R}^d$ captures the direction and speed of the agent's current movement. The settlement flag $a_{i,t} \in \{0,1\}$ is a binary indicator where $a_{i,t} = 1$ means that the agent has irreversibly settled, while $a_{i,t} = 0$ indicates that it remains free of motion.

The optimisation process is guided by a scalar objective function $f(x)$ that encodes the mission's exploration goals. The lower values of $f(x)$ correspond to more desirable locations characterised by higher environmental quality, lower operational risk, or greater information gain. The algorithm supports both single-objective and multi-objective formulations. A typical \ac{AUV} survey mission might employ an objective function (Equation~\ref{eq:objective_function}) of the form:
\begin{align}
f(x) &= w_1 \cdot \text{Energy}(x) + w_2 \cdot \text{Risk}(x) \nonumber \\
     &\quad - w_3 \cdot \text{Coverage}(x) - w_4 \cdot \text{InfoGain}(x) \nonumber \\
     &\quad + \text{Penalties},
\label{eq:objective_function}
\end{align}
which combines energy cost, risk exposure, coverage gaps, and information gain into a single score \cite{deb2011multi}. 

The environment is characterised by two additional components. The hydrodynamic flow field $U(x) \in \mathbb{R}^d$ represents the ambient ocean currents in position $x$, providing a natural drift bias for the agents. This can model either real ocean conditions or heuristic planning scenarios. The set of colonies $C_t$ contains all agents anchored at time $t$, where each colony $k \in C_t$ has a fixed location $c_k$ representing the settlement position where an agent was anchored, and a strength value $S_k$ indicating the influence of the colony and the communication capacity.

Colonies serve as local attractors and communication nodes for free agents. Initially, $C_0 = \emptyset$.

All spatial coordinates, distances, and fitness values are normalised to $[0,1]$ when appropriate to ensure consistent scaling throughout the algorithm. Once an agent settles (i.e. $a_{i,t}=1$), it ceases movement permanently. Formally, if agent $i$ settles at time $t$, then for all $t' > t$, its state remains fixed (Equation~\ref{eq:settlement_fixed}):
\begin{equation}
x_{i,t'} = x_{i,t}, \quad v_{i,t'} = \mathbf{0}.
\label{eq:settlement_fixed}
\end{equation}
This is enforced through the state update equations (Equation~\ref{eq:state_update}):
\begin{equation}
v_{i,t+1} = (1 - a_{i,t}) \,\tilde{v}_{i,t+1}, 
\quad
x_{i,t+1} = x_{i,t} + (1 - a_{i,t}) \,\tilde{v}_{i,t+1},
\label{eq:state_update}
\end{equation}
where $\tilde{v}_{i,t+1}$ is the tentative velocity update. The prefactor $(1 - a_{i,t})$ ensures settled agents remain frozen, implementing irreversible anchoring analogous to cyprid larvae \cite{kitade2022faint}. This progressive freeze strategy transitions agents from exploration to exploitation \cite{kirkpatrick1983optimization,cao2017optimization}, extending multi-swarm PSO frameworks \cite{liang2005dynamic} with irreversible anchoring.

\subsection{Drift and Update Equations}
At each iteration, every free agent ($a_{i,t}=0$) updates its velocity and position based on five components: inertia, random exploration, goal-seeking gradient, environmental drift, and colony influence. To capture inter-agent interactions, we introduce a novel colony field $\Phi(x)$ that models attraction and repulsion, and then present the overall update rule.

\paragraph{Colony influence field} 
Anchored colonies emit a spatial field that is repulsive in the short range (to avoid overcrowding) \cite{hooper2016spatial}, but attractive in the longer range (to draw agents into the vicinity). This bio-fouling pressure field mimics how barnacle colonies signal both attraction and exclusion zones, similar to how real barnacles use chemical cues to guide larval settlement \cite{Kotsiri2018}. The field incorporates communication noise through probabilistic signal transmission (Equation~\ref{eq:colony_field}), defined as
\begin{equation}
\begin{split}
\Phi(x) &= \sum_{k \in C_t} p_k(x) \left[
  A \exp\!\left(-\frac{\|x - c_k\|^2}{2\sigma_a^2}\right) \right. \\
&\quad \left. - B \exp\!\left(-\frac{\|x - c_k\|^2}{2\sigma_r^2}\right)
\right],
\end{split}
\label{eq:colony_field}
\end{equation}
where $p_k(x) \in [0,1]$ represents the probability of communication success between position $x$ and colony $k$ (Equation~\ref{eq:communication_probability}), modelled as:
\begin{equation}
\begin{split}
p_k(x) &= \mathbf{1}_{\|x - c_k\| \leq R_k} + \\
&\quad \mathbf{1}_{\|x - c_k\| > R_k} \exp(-\alpha(\|x - c_k\| - R_k))
\end{split}
\label{eq:communication_probability}
\end{equation}
where $A \in [0.1, 1.0]$ and $B \in [0.1, 0.5]$ control the attraction and repulsion strengths, while $\sigma_a \in [0.2, 0.8]$ and $\sigma_r \in [0.05, 0.3]$ define the spatial scales (with $\sigma_a > \sigma_r$). The communication parameters $R_k \in [0.1, 0.5]$ represent the reliable communication range for the colony $k$, and $\alpha \in [1.0, 5.0]$ controls signal attenuation beyond this range, reflecting realistic underwater acoustic propagation characteristics \cite{connor2021current}. 

The parameter ranges reflect biological observations. Attraction strength $A$ corresponds to chemical signalling intensity in barnacle colonies, while repulsion strength $B$ models physical crowding constraints. The spatial scales $\sigma_a$ and $\sigma_r$ mirror the effective range of chemical signals versus the zones of immediate physical contact observed in marine larvae \cite{wright1999influence}. 

Based on cyprid swimming speeds of $\sim$1.8 cm/s \cite{larsson2016instantaneous, Kotsiri2018}, the attraction ranges significantly exceed the immediate contact zones. This enables larvae to detect and navigate the appropriate settlement sites from considerable distances. The dual-scale approach ensures that agents are drawn toward productive areas, but avoid overcrowding. This mirrors the biological phenomenon where barnacle larvae are attracted to established colonies but settle at appropriate distances to avoid competition for resources. 

The gradient $G_{\Phi}(x)=\nabla\Phi(x)$ acts as a virtual force on free agents, guiding their movement toward optimal settlement regions while maintaining spatial diversity \cite{gazi2004class, gazi2006coordination}. This dual-scale interaction mirrors empirical observations in barnacle larvae, where cyprids are chemically attracted to productive conspecific aggregations, but still reject immediate-contact regions to avoid overcrowding \cite{Berntsson2004}. Furthermore, larvae can detect dissolved settlement signals under flow before making direct contact with the substrate, allowing field-driven navigation rather than purely contact-based adhesion \cite{Koehl2004}. The Gaussian-based potential field approach follows established methods of the artificial potential field for swarm robotics \cite{baziyad2023direction}, while the communication probability model incorporates realistic characteristics of the wireless link observed in transition regions \cite{zuniga2004analyzing}. The attraction-repulsion dynamics are based on social force models \cite{helbing1995social} (see Equation \ref{eq:colony_field}).

\paragraph{Velocity update}
The velocity update mechanism integrates five biologically inspired components that mirror how barnacle nauplii navigate within hydrodynamic environments. Let $g(x_{i,t})$ denote a surrogate estimate of the local gradient of $f$ at the position of agent $i$. This estimate may be obtained through finite differences or relative fitness ranks while keeping NOAH derivative-free. The tentative velocity update $\tilde{v}_{i,t+1}$ is given by:
\begin{align}
\tilde{v}_{i,t+1} &= \underbrace{\omega v_{i,t}}_{\text{inertia}} + \underbrace{\eta \xi_{i,t}}_{\text{random}} \nonumber \\
&\quad + \underbrace{\beta \left(-g(x_{i,t})\right)}_{\text{fitness gradient}} + \underbrace{\gamma U(x_{i,t})}_{\text{drift (flow)}} \nonumber \\
&\quad + \underbrace{\delta G_{\Phi}(x_{i,t})}_{\text{colony field}},
\label{eq:velocity_update}
\end{align}
where $\xi_{i,t} \sim \mathcal{N}(\mathbf{0},\mathbf{I})$ introduces stochastic exploration. The inertia term $(\omega v_{i,t})$ follows the particle swarm optimisation framework \cite{shi1998modified}, where the inertia weight $\omega$ balances global and local search capabilities \cite{shi2001comparing}. The random component $(\eta\xi_{i,t})$ employs Gaussian perturbations consistent with stochastic approximation theory, particularly simultaneous perturbation stochastic approximation methods \cite{spall1999adaptive,maryak2001global}. The fitness gradient term $(\beta (-\hat{\nabla} f))$ uses finite difference approximations for derivative-free optimisation \cite{boresta2022mixed,khanh2023general}, while the drift term $(\gamma U(x_{i,t}))$ models the hydrodynamic flow effects observed in ocean sampling applications \cite{paley2008cooperative}. Finally, the colony field $(\delta G_{\Phi}(x_{i,t}))$ incorporates the mean-field control principles for swarm coordination \cite{carra2017coalescing,cui2024density}.

The parameter ranges are biologically motivated based on observed swimming speeds of 2.6 to 15.2 body lengths per second \cite{wong2020swimming} and rheotactic attachment behaviours in which 93\% of cyprids swim upstream before settlement \cite{larsson2016instantaneous}. Momentum retention $\omega \in [0.5, 0.9]$ reflects larval inertia, exploration parameter $\eta \in [0.1, 0.5]$ models stochastic swimming variability, gradient following strength $\beta \in [0.2, 1.0]$ corresponds to chemotactic responsiveness, current adaptation $\gamma \in [0.3, 0.8]$ reflects hydrodynamic flow exploitation, and colony influence $\delta \in [0.1, 0.6]$ balances attraction to settlement-rich regions with localised spacing effects \cite{carrillo2015living}.

\paragraph{Autonomous hydrodynamics}
The drift term $\gamma U(x_{i,t})$ represents the core capability of \ac{NOAH} for autonomous hydrodynamics, enabling agents to exploit ocean currents rather than resist them. Unlike traditional swarm algorithms that treat environmental forces as disturbances that must be overcome, \ac{NOAH} agents actively use the hydrodynamic flow field $U(x)$ to improve exploration efficiency and reduce energy consumption. When $U(x_{i,t})$ points toward promising regions, agents accelerate their convergence; when currents flow away from optimal areas, agents can still navigate effectively through the combination of gradient following and colony attraction. This autonomous adaptation to hydrodynamic conditions mirrors how barnacle nauplii exploit favourable flow structures for settlement \cite{larsson2016instantaneous}, transforming environmental constraints into navigational advantages. The parameter $\gamma \in [0.3, 0.8]$ controls the degree of current exploitation, allowing \ac{NOAH} to operate effectively in diverse marine environments, from gentle coastal flows to strong oceanic currents.

\paragraph{State update.}
After computing $\tilde{v}_{i,t+1}$, the agent’s motion is updated as
\begin{equation}
v_{i,t+1} = (1-a_{i,t})\,\tilde{v}_{i,t+1}, \quad
x_{i,t+1} = x_{i,t} + (1-a_{i,t})\,\tilde{v}_{i,t+1},
\label{eq:final_state_update}
\end{equation}
ensuring that the settled agents ($a_{i,t}=1$) remain fixed. Any $x_{i,t+1}$ that falls outside the search domain (e.g. due to drift) is projected back into $[0,1]^d$, using reflecting or wrapping boundary conditions as appropriate for the application.

\subsection{Settlement Decision and Irreversibility}
Each free agent autonomously decides whether to settle at its current location using a probabilistic model that incorporates five key features: local fitness advantage, colony distance, flow stability, crowding, and energy state. If an agent settles, it becomes irreversibly anchored. The probability of settlement is calculated using a logistic sigmoid function $\sigma(x) = \frac{1}{1 + e^{-x}}$:

\begin{align}
p_{i,t}^{\text{settle}} &= \sigma\!\left(
\lambda_1 \,\text{rank}(\Delta f_i) + \lambda_2 \frac{d_i}{d_0} \right. \nonumber \\
&\quad \left. - \lambda_3 \frac{\kappa(x_{i,t})}{\kappa_0} - \lambda_4 \rho_i + \lambda_5 E_i
\right),
\label{eq:settlement_probability}
\end{align}

where $\Delta f_i = \tilde{f}_N - f(x_{i,t})$ represents the local fitness advantage, $d_i = \min_{k\in C_t} \|x_{i,t} - c_k\|$ is the distance to the nearest colony, $\kappa(x_{i,t}) = \|\nabla U(x_{i,t})\|_2$ measures flow shear (turbulence), $\rho_i$ indicates local crowding and $E_i \in [0,1]$ encodes energy reserves. The parameters $\lambda_1,\dots,\lambda_5$ control the relative importance of each factor, while $d_0$ and $\kappa_0$ are normalising constants. Here, $\text{rank}(\Delta f_i)$ denotes the scalar rank of agent $i$ among all agents (1 for best fitness, 2 for second-best, etc.), not the matrix rank.

The key terms in the settlement probability are defined as follows. The average fitness of the neighbourhood $\tilde{f}_N$ (Equation~\ref{eq:neighborhood_fitness}) represents the mean fitness of agents within a local radius $r_{\text{neigh}}$ of the agent $i$:
\begin{equation}
\tilde{f}_N = \frac{1}{|N_i|} \sum_{j \in N_i} f(x_{j,t}), \quad N_i = \{j : \|x_{i,t} - x_{j,t}\| \leq r_{\text{neigh}}\},
\label{eq:neighborhood_fitness}
\end{equation}
where $N_i$ is the neighbourhood set of agent $i$. Local crowding $\rho_i$ (Equation~\ref{eq:local_crowding}) quantifies the agent density within the settlement radius $r_{\text{settle}}$:
\begin{equation}
\rho_i = \frac{|S_i|}{V_{\text{settle}}}, \quad S_i = \{j : \|x_{i,t} - x_{j,t}\| \leq r_{\text{settle}}\},
\label{eq:local_crowding}
\end{equation}
where $V_{\text{settle}} = \pi r_{\text{settle}}^2$ is the volume of the settlement area in 2D (or $V_{\text{settle}} = \frac{4}{3}\pi r_{\text{settle}}^3$ in 3D). The energy state $E_i$ (Equation~\ref{eq:energy_state}) represents the normalised remaining energy capacity of agent $i$, computed as:
\begin{equation}
E_i = \frac{E_{\text{current},i}}{E_{\text{max},i}},
\label{eq:energy_state}
\end{equation}
where $E_{\text{current},i}$ is the current energy level and $E_{\text{max},i}$ is the maximum energy capacity of the agent $i$.

After computing the probability of settlement, each agent samples $u \sim U(0,1)$ and anchors if $u < p_{i,t}^{\text{settle}}$, creating a new colony with initial strength $S_{\text{new}}$. This decision is irreversible, as once an agent settles, it becomes a permanent sensor node, analogous to barnacle cyprid metamorphosis.

\subsection{Colony Dynamics and Communication}
Once agents begin to settle, the colony system becomes active, creating a dynamic communication and influence network. When a new colony is spawned by an agent that settles, it is added to the colony set $C_t$ with an initial strength reflecting the quality of the site. For a new colony at $x_{i,t}$ we set
\begin{equation}
S_{\text{new}} = \alpha_0 + \alpha_1\,\text{rank}\!\left(-f(x_{i,t})\right),
\label{eq:colony_initial_strength}
\end{equation}
where $\text{rank}(-f(x_{i,t}))$ is the relative rank of the fitness of the agent among the population before settlement, and $\alpha_0,\alpha_1$ are constants. The ranking function assigns integer ranks in ascending order of fitness values, where the best (lowest) fitness receives rank 1, the second-best receives rank 2, etc. Thus, every new colony begins with a base strength $\alpha_0$ and receives an additional boost if the agent had a particularly low objective value (i.e. high quality discovery). This ensures that colonies established by high-performing agents exert a stronger initial influence, analogous to how a resource-rich site immediately attracts settlers.

\paragraph{Colony strength update.} 
Over time, colony strength $S_k$ is dynamically adjusted based on reinforcement from local successes and penalties from overcrowding, subject to the probability of communication success. Let $\overline{\text{rank}(-f)}_{\text{near }c_k}$ denote the average of these scalar ranks for agents near colony $k$, and let $\overline{\rho}_{\text{near }c_k}$ be the average agent density in its vicinity. The update rule incorporates communication probability as
\begin{align}
S_k &\leftarrow (1-\mu)S_k + \mu \cdot p_k^{\text{comm}} \!\left(\overline{\text{rank}(-f)}_{\text{near }c_k} \right. \nonumber \\
&\quad \left. - \nu\,\overline{\rho}_{\text{near }c_k}\right),
\label{eq:colony_strength_update}
\end{align}
where $\mu\in(0,1]$ is a learning rate and $\nu$ scales the crowding penalty. The probability of communication success $p_k^{\text{comm}} \in [0,1]$ is calculated as the average of $p_k(x)$ on all nearby agents, representing the reliability of communication between the colony $k$ and its surrounding agents. When $p_k^{\text{comm}} = 1$, the colony receives full updates; when $p_k^{\text{comm}} = 0$, the colony strength decays without reinforcement, simulating communication failure. This update increases colony strength if nearby agents continue to produce high-quality solutions and the area is not over-saturated. In contrast, if local performance declines or density grows excessively, strength decreases. Colonies thus behave as living entities: reinforced by successful discoveries and weakened by redundancy. This mechanism parallels the reinforcement of pheromones and evaporation in the optimisation of the ant colony \cite{dorigo2006ant}.

\paragraph{Colony culling.}
If the strength of a colony falls below a threshold $\tau$ ($S_k < \tau$), it is removed from the active colony set $C_t$. The anchored AUV remains physically fixed, but ceases to emit influence or participate in communication. This prevents obsolete colonies from misleading the swarm, akin to barnacles in poor locations that die or lose reproductive signalling capacity while their shells remain settled.

\paragraph{Colony-based communication.}
The set of active colonies forms a communication backbone for the swarm. Biologically, barnacles induce settlement via cues such as the settlement-inducing protein complex (SIPC) deposited on surfaces and waterborne pheromones (e.g., adenosine). In NOAH, this analogy is captured through the colony field $\Phi(x)$ and its gradient $G_{\Phi}(x)$. 

An anchored AUV can transmit acoustic or optical beacons. These function both as attractors ("this site is valuable") and as exclusion sources that prevent overcrowding. Through $G_{\Phi}(x)$, free agents implicitly sense colony locations and strengths. This leads to clustering around productive regions while maintaining the spacing between colonies.

Beyond settlement cues, anchored AUVs can act as stationary relay nodes for acoustic networking. By serving as reliable communication hubs, colonies enable multi-hop data exchange across the swarm. This improves robustness and coverage. The emergent backbone resembles the way barnacle colonies collectively enhance feeding and reproduction. It has parallels in AUV relay deployment strategies that leverage anchored robots to extend communication range.

\subsection{Complexity and Convergence Considerations}

The computational complexity introduces additional overhead compared to standard swarm optimisers such as PSO \cite{kennedy1995particle} and ACO \cite{dorigo2006ant}. This is due to the colony field computation and settlement decision processes. Let $N$ denote the number of agents and $C$ the number of active colonies at time $t$. 

Each iteration requires $O(N)$ operations for agent state updates, $O(N \cdot C)$ for colony field computation $G_{\Phi}(x)$ at each agent location, $O(N)$ for settlement probability evaluation using the logistic model \cite{fagerland2012generalized}, and $O(C)$ for colony strength updates based on pheromone reinforcement mechanisms \cite{dorigo2006ant}. In the worst case where $C = O(N)$ (all agents eventually settle), the total complexity per iteration is $O(N^2)$. 

In practice, the complexity of the average-case is significantly better. During the initial exploration phase, $C \ll N$, resulting in $O(N)$ complexity per iteration. As agents begin to settle, $C$ gradually grows, but typically remains $C \leq \sqrt{N}$ in most scenarios, yielding a mean case complexity of $O(N^{1.5})$ per iteration. Memory complexity is $O(N)$ since the algorithm requires storage for agent states and colony information where $C \leq N$.

The communication architecture is specifically designed to minimise bandwidth requirements in underwater acoustic networks, where communication is expensive and unreliable \cite{connor2021current}. Unlike approaches where all agents continuously exchange information, the algorithm leverages colonies to reduce communication load through a hierarchical structure. Free agents broadcast locally (positions, fitness values) and listen for colony beacons, while anchored agents act as cluster heads, broadcasting colony strength and location information. During iteration, the complexity of communication is $O(N + C)$ messages, with $O(N)$ local broadcasts from free agents and $O(C)$ colony broadcasts. This represents a significant reduction compared to fully-connected swarm algorithms and enables multi-hop acoustic networking through anchored agents.

The convergence properties are fundamentally different from traditional swarm algorithms because of the irreversible anchoring mechanism. This creates a progressive transition from exploration to exploitation reminiscent of simulated annealing. The number of free agents $N_{\text{free}}(t)$ decreases monotonically over time when the agents settle, since $a_{i,t} = 1$ is irreversible. Under proper parameter tuning, the algorithm converges when $N_{\text{free}}(t) = 0$, resulting in a static configuration of colonies $C$ colonies.

Anchoring is implemented as a monotonic freeze strategy, akin to absorbing states in Markov decision processes: once an agent commits, it becomes an absorbing node and cannot re-enter the mobile state. This guarantees convergence in finite time, since the number of free agents is strictly decreasing. 

The algorithm implements a progressive freeze strategy. It begins with an initial exploration phase where all agents explore the search space using velocity updates that incorporate inertia \cite{shi1998modified}, random exploration with Gaussian noise \cite{wang2003stochastic} and gradient estimation \cite{zhang2012local}. During the discovery phase, high-quality regions are identified and secured through anchoring. The exploitation phase guides the remaining free agents to promising areas through colony fields based on attraction-repulsion dynamics \cite{gazi2004class} \cite{gazi2006coordination}. The system converges to a metastable state where all agents are anchored, creating a distributed sensor network.

While formal global optimality proofs are challenging due to the algorithm's stochastic nature, several convergence properties can be established. The algorithm terminates in finite time with high probability when settlement parameters are properly tuned, providing practical convergence guarantees for underwater exploration missions. Colony field repulsion is designed to prevent clustering at single optima, potentially enabling multimodal coverage essential for underwater exploration missions. Anchored colonies preserve discovered optima from being lost, maintaining quality preservation throughout the optimisation process. The quality of convergence depends on the parameter tuning, where high values of $\lambda_1$ may cause premature anchoring in local optima, while low settlement thresholds encourage exploration but delay convergence. The parameters of the colony field ($A$, $B$, $\sigma_a$, $\sigma_r$) balance the dynamics of attraction and repulsion.

\subsection{Flowchart Representation}

\begin{figure}[htbp]
    \centering
    \includegraphics[width=1\columnwidth,height=1\textheight,keepaspectratio]{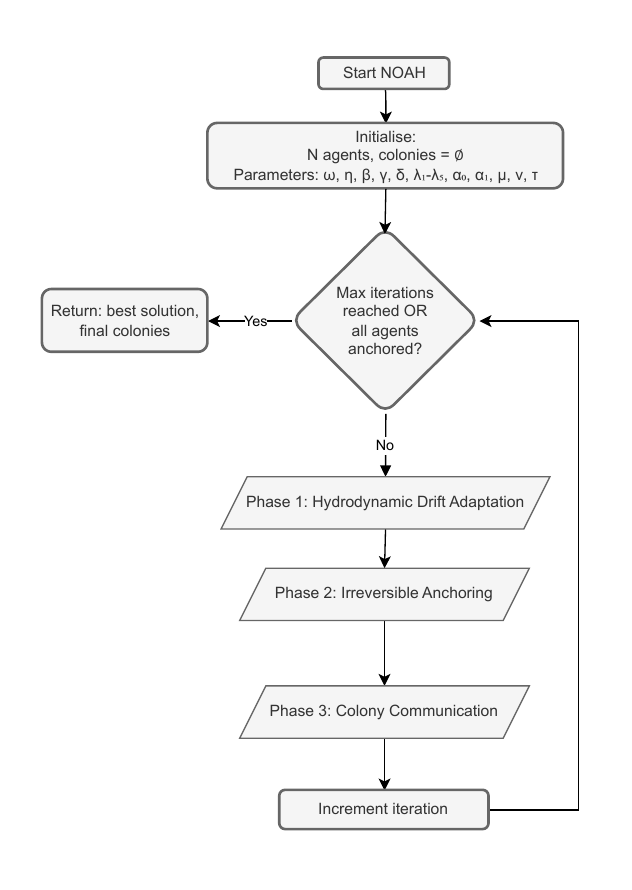}
    \caption{\ac{NOAH} algorithm workflow showing the integration of optimisation phases from initialisation to convergence.}
    \label{fig:noah_flowchart}
\end{figure}

\label{subsec:flowchart_representation}

Fig.~\ref{fig:noah_flowchart} provides a visual overview of the workflow of the \ac{NOAH} algorithm, illustrating the integration of the three core phases: Hydrodynamic Drift Adaptation, Irreversible Anchoring, and Colony Communication. The flowchart shows the complete process from initialisation through convergence, where all agents transition from free exploration to anchored sensor nodes.

\subsection{Pseudo Code Representation}
\label{subsec:pseudo_code}

The algorithm \ac{NOAH} is implemented through four interconnected algorithms (Algorithm~\ref{alg:noah_complete}) that provide the complete algorithmic specification with clear start conditions, execution flow, and exit criteria. The algorithm requires an objective function $f$, a flow field $U$, parameters ($\omega, \eta, \beta, \gamma, \delta, \lambda_1, \lambda_2, \lambda_3, \lambda_4, \lambda_5, \alpha_0, \alpha_1, \mu, \nu, \tau$), dimensions $d$, and maximum iterations. The output consists of optimisation results with the best solution found and the final colony configuration. A complete description of all mathematical notation is provided in Table~\ref{tab:notation}.

\subsubsection{Hydrodynamic Drift Adaptation}
\label{subsubsec:phase1}

The first phase deals with the adaptation of hydrodynamic drift, where all agents begin as free ($a_{i,t} = 0$). During this phase, free agents evaluate their fitness, update velocities using Equation~\ref{eq:velocity_update}, and update positions using Equation~\ref{eq:final_state_update}. This phase ends when all free agents have moved to new positions based on inertia, random exploration, gradient guidance, hydrodynamic drift, and colony field influence.

\begin{algorithm}
\caption{Phase 1: Hydrodynamic Drift Adaptation}
\label{alg:noah_phase1}
\begin{algorithmic}[1]
\STATE \textbf{Input:} agents, flow\_field $U$, parameters ($\omega, \eta, \beta, \gamma, \delta$)
\STATE \textbf{Output:} updated\_agents, best\_solution, best\_fitness
\FOR{each agent $i$ in agents}
    \IF{agent.settlement\_flag $= 0$}
        \STATE \textbf{Evaluate fitness:} $f_i = f(x_{i,t})$
        \STATE \textbf{Update best solution:} 
        \IF{$f_i <$ best\_fitness}
            \STATE best\_solution $= x_{i,t}$, best\_fitness $= f_i$
        \ENDIF
        \STATE \textbf{Update velocity using Equation~\ref{eq:velocity_update}:}
        \STATE $\tilde{v}_{i,t+1} = \omega v_{i,t} + \eta \xi_{i,t} + \beta(-g(x_{i,t})) + \gamma U(x_{i,t}) + \delta G_{\Phi}(x_{i,t})$
        \STATE \textbf{Update position using Equation~\ref{eq:final_state_update}:}
        \STATE $v_{i,t+1} = (1-a_{i,t})\tilde{v}_{i,t+1}$
        \STATE $x_{i,t+1} = x_{i,t} + (1-a_{i,t})\tilde{v}_{i,t+1}$
    \ENDIF
\ENDFOR
\RETURN \{updated\_agents, best\_solution, best\_fitness\}
\end{algorithmic}
\end{algorithm}

\subsubsection{Irreversible Anchoring}
\label{subsubsec:phase2}

The second phase manages irreversible anchoring, beginning with all free agents evaluating their settlement probability using Equation~\ref{eq:settlement_probability}. Agents probabilistically decide to anchor based on local fitness advantage, colony distance, flow stability, crowding, and energy state. This phase ends when the anchored agents become stationary colonies with initial strength computed from Equation~\ref{eq:colony_initial_strength}.

\begin{algorithm}
\caption{Phase 2: Irreversible Anchoring}
\label{alg:noah_phase2}
\begin{algorithmic}[1]
\STATE \textbf{Input:} agents, colonies, parameters ($\lambda_1, \lambda_2, \lambda_3, \lambda_4, \lambda_5, \alpha_0, \alpha_1$)
\STATE \textbf{Output:} updated\_agents, new\_colonies
\FOR{each agent $i$ in agents}
    \IF{agent.settlement\_flag $= 0$}
        \STATE \textbf{Compute settlement probability using Equation~\ref{eq:settlement_probability}:}
        \STATE $p_{i,t}^{\text{settle}} = \sigma(\lambda_1 \text{rank}(\Delta f_i) + \lambda_2 \frac{d_i}{d_0} - \lambda_3 \frac{\kappa(x_{i,t})}{\kappa_0} - \lambda_4 \rho_i + \lambda_5 E_i)$
        \STATE \textbf{Sample and anchor:}
        \IF{$u \sim U(0,1) < p_{i,t}^{\text{settle}}$}
            \STATE agent.settlement\_flag $= 1$
            \STATE \textbf{Create colony with initial strength using Equation~\ref{eq:colony_initial_strength}:}
            \STATE $S_{\text{new}} = \alpha_0 + \alpha_1 \cdot \text{rank}(-f(x_{i,t}))$
            \STATE new\_colony $=$ CreateColony($x_{i,t}$, $S_{\text{new}}$)
            \STATE new\_colonies $=$ new\_colonies $\cup$ \{new\_colony\}
        \ENDIF
    \ENDIF
\ENDFOR
\RETURN \{updated\_agents, new\_colonies\}
\end{algorithmic}
\end{algorithm}

\subsubsection{Colony Communication}
\label{subsubsec:phase3}

The third phase handles colony communication, starting when active colonies exist in the environment. The strength of the colonies is updated using Equation~\ref{eq:colony_strength_update}, and weak colonies ($S_k < \tau$) are culled while remaining stationary. This phase ends when the colony network is optimised for communication and influence.

\begin{algorithm}
\caption{Phase 3: Colony Communication}
\label{alg:noah_phase3}
\begin{algorithmic}[1]
\STATE \textbf{Input:} colonies, agents, parameters ($\mu, \nu, \tau$)
\STATE \textbf{Output:} updated\_colonies, active\_colonies
\FOR{each colony $k$ in colonies}
    \STATE \textbf{Compute communication probability:} $p_k^{\text{comm}} =$ AverageCommunicationSuccess($c_k$, nearby\_agents)
    \STATE \textbf{Update colony strength using Equation~\ref{eq:colony_strength_update}:}
    \STATE $S_k = (1-\mu)S_k + \mu \cdot p_k^{\text{comm}} \cdot (\overline{\text{rank}(-f)}_{\text{near }c_k} - \nu \cdot \overline{\rho}_{\text{near }c_k})$
    \STATE \textbf{Note:} Colony $k$ remains fixed at position $c_k$ (irreversible anchoring)
\ENDFOR
\STATE \textbf{Cull weak colonies:} Remove colonies where $S_k < \tau$ (anchored agents remain stationary)
\STATE active\_colonies $=$ \{colonies where $S_k \geq \tau$\}
\RETURN \{updated\_colonies, active\_colonies\}
\end{algorithmic}
\end{algorithm}

\subsubsection{Algorithm Integration and Execution}
\label{subsubsec:complete_algorithm}

The complete algorithm \ac{NOAH} (Algorithm~\ref{alg:noah_complete}) orchestrates these three phases sequentially until convergence is achieved. The algorithm ends when either all agents are anchored ($N_{\text{free}}(t) = 0$) or maximum iterations are reached. Convergence occurs when the progressive freeze strategy is complete, creating a distributed sensor network.

\begin{algorithm}
\caption{Complete \ac{NOAH} Algorithm}
\label{alg:noah_complete}
\begin{algorithmic}[1]
\STATE \textbf{Initialise:} agents $=$ InitialiseSwarm($N$), colonies $=$ $\emptyset$, iteration $=$ 0
\STATE \textbf{Initialise optimisation:} best\_solution $=$ $\emptyset$, best\_fitness $=$ $\infty$
\FOR{iteration $= 1$ to max\_iterations}
    \STATE \textbf{Execute Phase 1:} \{agents, best\_solution, best\_fitness\} $=$ Algorithm~\ref{alg:noah_phase1}
    \STATE \textbf{Execute Phase 2:} \{agents, new\_colonies\} $=$ Algorithm~\ref{alg:noah_phase2}
    \STATE colonies $=$ colonies $\cup$ new\_colonies
    \STATE \textbf{Execute Phase 3:} \{colonies, active\_colonies\} $=$ Algorithm~\ref{alg:noah_phase3}
    \STATE colonies $=$ active\_colonies
    \STATE \textbf{Check convergence:} free\_agents $=$ CountFreeAgents(agents)
    \IF{free\_agents $= 0$}
        \STATE \textbf{BREAK} \COMMENT{Algorithm converged - all agents anchored}
    \ENDIF
\ENDFOR
\RETURN \{best\_solution, best\_fitness, colonies, final\_colony\_count\}
\end{algorithmic}
\end{algorithm}

\section{Validation and Performance Evaluation}
\label{sec:validation}

Validation of \ac{NOAH} requires a multi-faceted approach that evaluates both its unique irreversible anchoring mechanism and its performance relative to established optimisation methods. This section presents three complementary validation studies: anchor accuracy validation, convergence behaviour analysis (Fig.~\ref{fig:convergence}), and comparative performance evaluation against baseline methods (Fig.~\ref{fig:ranking}). These studies collectively demonstrate \ac{NOAH}'s effectiveness for underwater robotic deployment scenarios.

\subsection{Anchoring Accuracy Validation}
\label{subsec:anchoring_validation}

\begin{figure*}[htbp]
    \centering
    \includegraphics[width=0.8\textwidth]{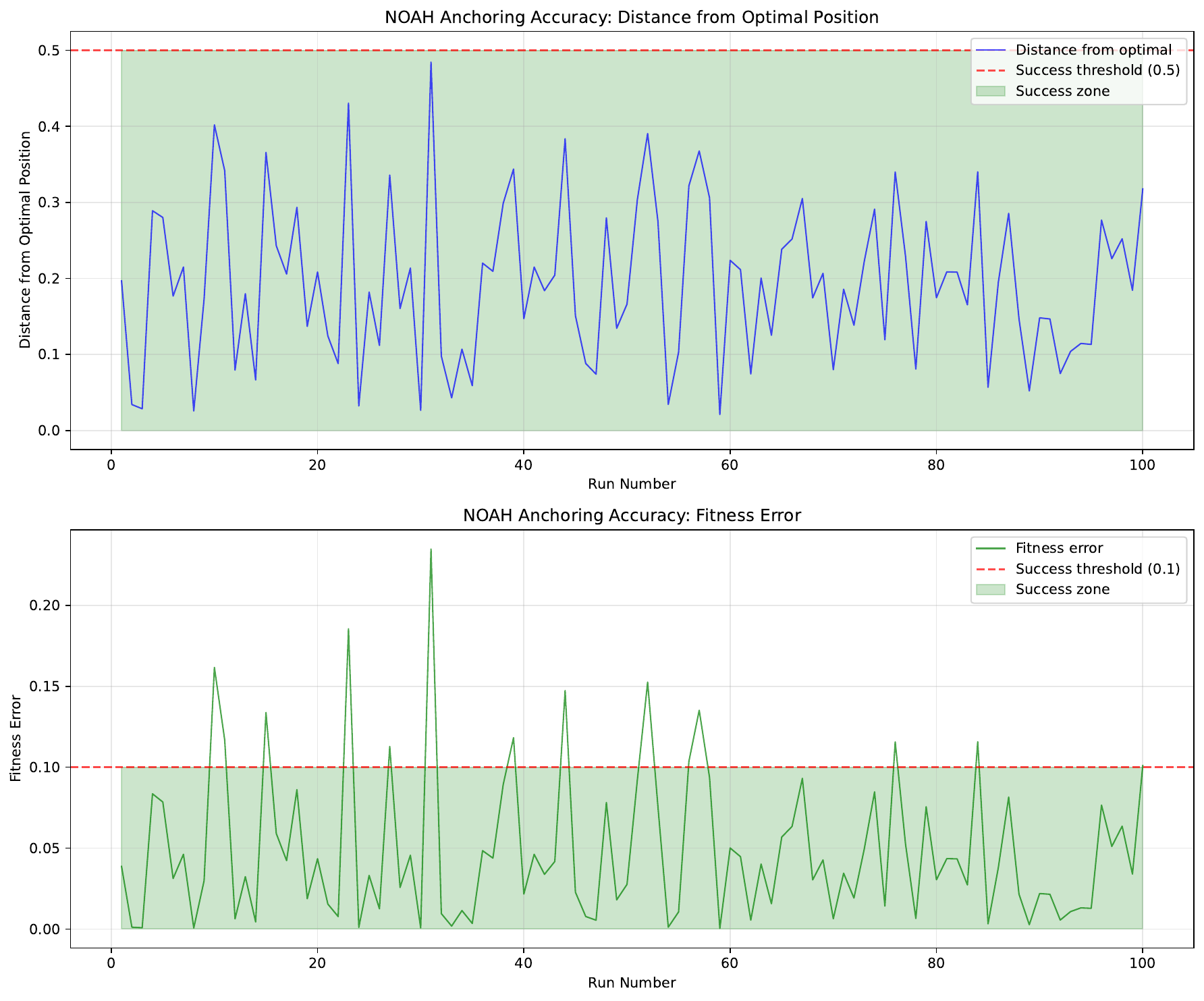}
    \caption{NOAH anchoring accuracy across 100 independent runs. Green markers indicate successful anchoring (within 0.5 units of optimum), red markers indicate failed attempts. The distribution demonstrates \ac{NOAH}'s ability to identify and anchor at optimal locations under permanent settlement constraints.}
    \label{fig:anchoring_accuracy}
\end{figure*}

\begin{table}[htbp]
\centering
\caption{NOAH Algorithm Parameter Configuration}
\label{tab:noah_parameters}
\begin{tabular}{@{}lll@{}}
\toprule
\textbf{Parameter} & \textbf{Value} & \textbf{Description} \\
\midrule
\multicolumn{3}{l}{\textbf{Core Movement Parameters}} \\
$\omega$ & 0.9 & Inertia weight \\
$\eta$ & 0.3 & Exploration factor \\
$\beta$ & 0.8 & Gradient following \\
$\gamma$ & 0.6 & Current adaptation \\
$\delta$ & 0.4 & Colony influence \\
\midrule
\multicolumn{3}{l}{\textbf{Settlement Parameters}} \\
$\lambda_1$ & 2.0 & Fitness advantage weight \\
$\lambda_2$ & 1.0 & Distance weight \\
$\lambda_3$ & 0.5 & Flow stability weight \\
$\lambda_4$ & 1.0 & Crowding weight \\
$\lambda_5$ & 1.0 & Energy weight \\
$\tau$ & 0.1 & Colony threshold \\
$\alpha_0$ & 1.0 & Base colony strength \\
$\alpha_1$ & 0.5 & Fitness-based strength \\
\bottomrule
\end{tabular}
\end{table}

The anchor accuracy test evaluates \ac{NOAH}'s fundamental ability to identify and permanently anchor in optimal locations. Using a custom underwater test function $f(x,y) = \|[x,y]\|^2 - 1$ with minimum $-1$ at $(0,0)$, we assess whether agents can locate and commit to optimal deployment sites under irreversible settlement constraints. The test environment comprised a 2D search space $[-2.0, 2.0]^2$, 20 agents, 50 iterations, circular flow pattern, and 100 independent runs. The algorithm parameter configuration is detailed in Table~\ref{tab:noah_parameters}.

Results demonstrate 86\% success (86/100 runs). Distance from optimum ranged from 0.021 to 0.485 units (mean: 0.194±0.104), with fitness error ranging from 0.0004 to 0.235 (mean: 0.049±0.046). Best performance achieved position $(-0.020, 0.008)$ with fitness $-0.9996$, demonstrating near-perfect accuracy. Colony formation averaged 5.11 per run, validating reliability for permanent deployment.

\subsection{Convergence Behaviour Analysis}
\label{subsec:convergence_analysis}

\begin{figure}[htbp]
    \centering
    \includegraphics[width=\columnwidth]{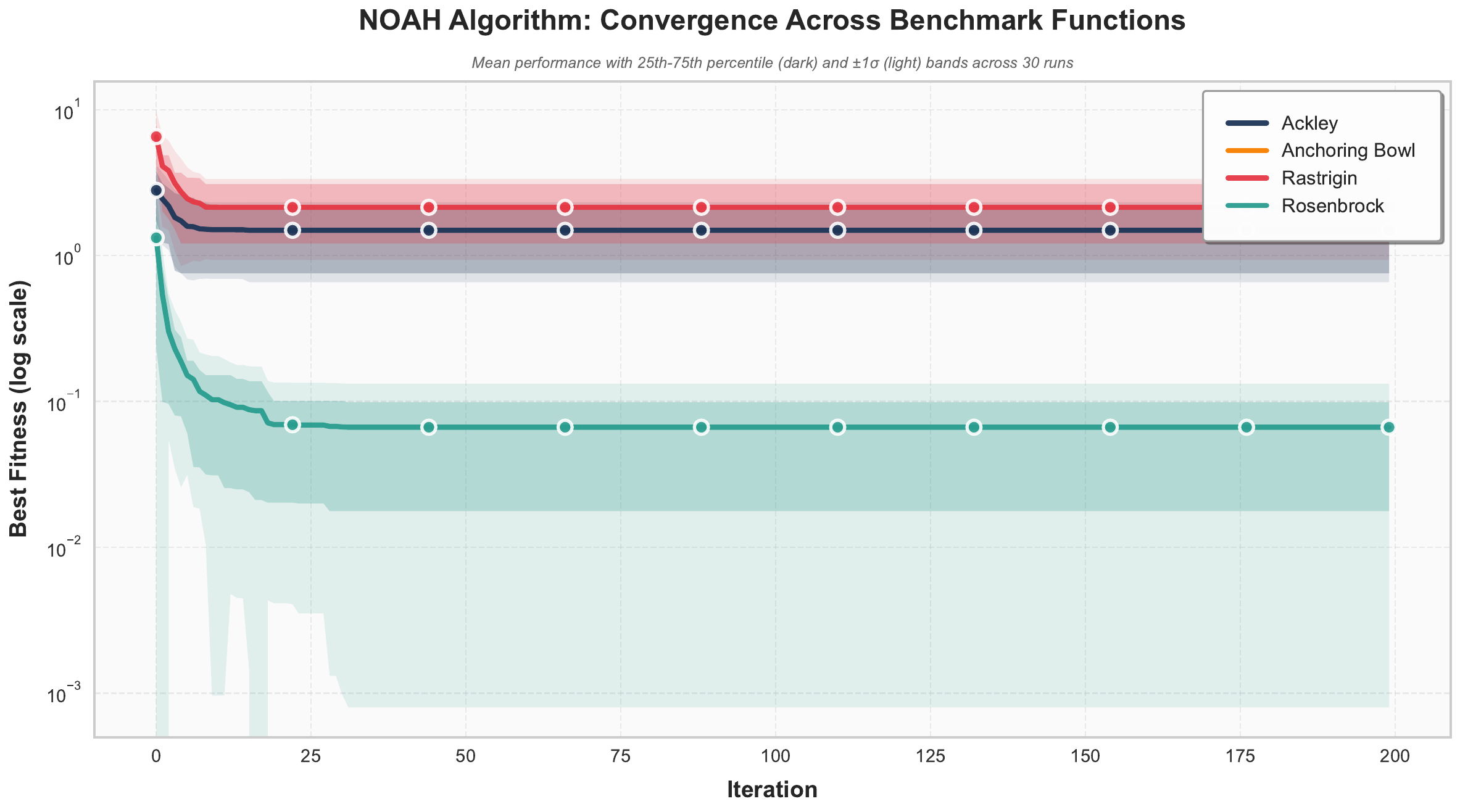}
\caption{NOAH convergence behaviour across four benchmark functions over 200 iterations. Solid lines represent mean best fitness, shaded regions indicate standard deviation across 30 independent runs. The progressive improvement demonstrates effective exploration-exploitation balance through irreversible anchoring.}
    \label{fig:convergence}
\end{figure}

The convergence analysis evaluates \ac{NOAH}'s optimisation dynamics across diverse problem landscapes. Four benchmark functions assess different algorithmic capabilities: Rastrigin (highly multimodal with many local optima), Ackley (multimodal with deep global minimum), Rosenbrock (unimodal with narrow valley), and Anchoring Bowl (custom bowl-shaped function simulating underwater deployment scenarios). The experimental configuration employed 50 agents, 200 iterations, and 30 random seeds per function to ensure statistical significance.

Figure~\ref{fig:convergence} shows convergence patterns: Rastrigin $2.14 \pm 1.21$ (67.2\% improvement), Ackley $1.49 \pm 0.84$ (46.8\%), Rosenbrock $0.067 \pm 0.066$ (95.0\%), and Anchoring Bowl $-0.991 \pm 0.010$ (near-optimal).

All functions exhibited rapid initial improvement within 20-30 iterations, followed by progressive refinement via irreversible anchoring. Confidence bands indicate consistent performance across initialisations, validating algorithmic robustness.

\subsection{Comparative Performance Evaluation}
\label{subsec:comparative_evaluation}

\begin{figure}[htbp]
    \centering
    \includegraphics[width=\columnwidth]{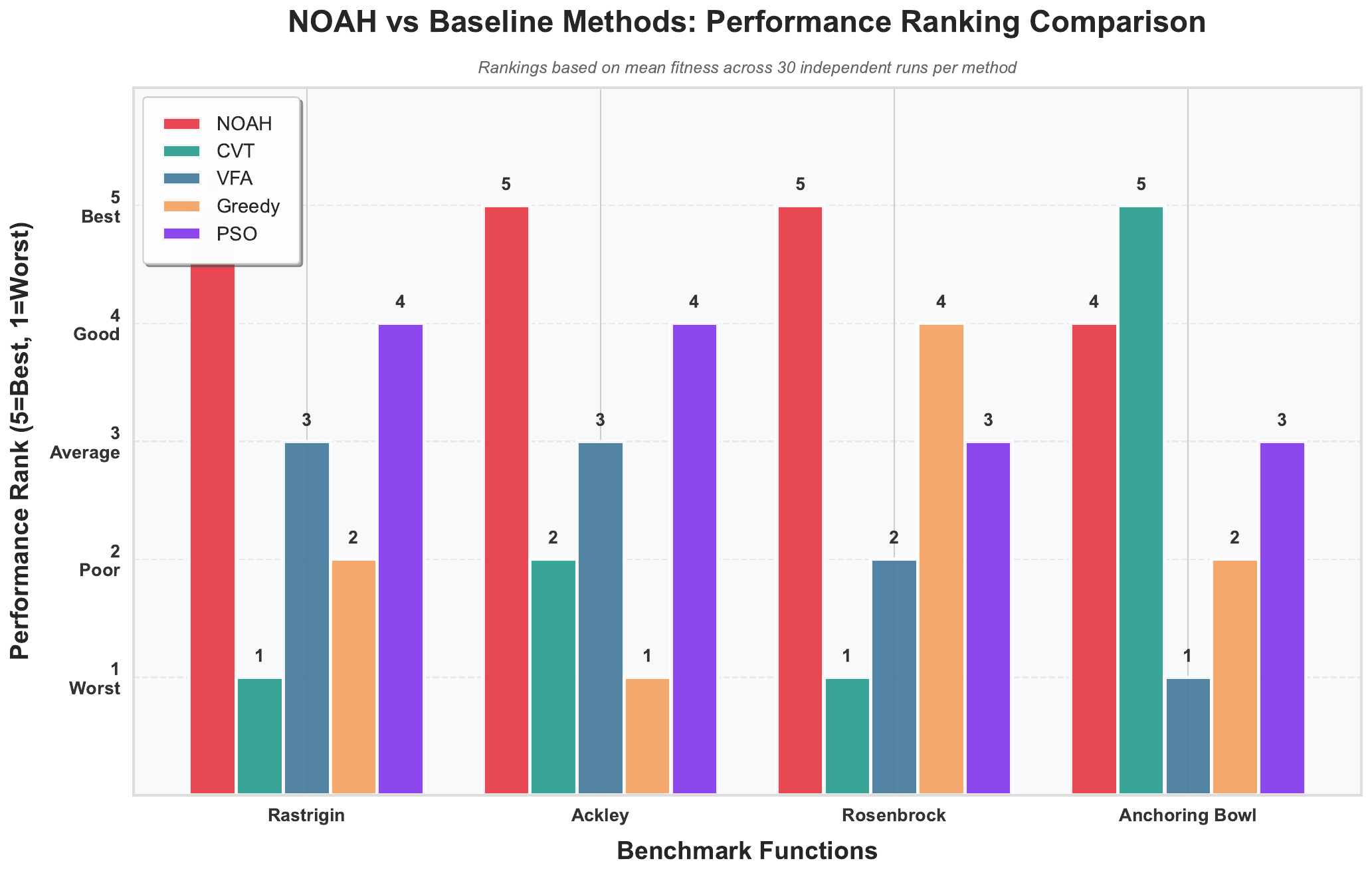}
    \caption{Performance ranking heatmap comparing NOAH against four baseline methods across benchmark functions. Rankings range from 1 (worst) to 5 (best) based on mean best fitness. NOAH achieves first place in 3 of 4 benchmarks with average ranking 4.75/5.0.}
    \label{fig:ranking}
\end{figure}

Comparative evaluation benchmarks \ac{NOAH} against four established methods: \ac{PSO} (mobile swarm reference), \ac{CVT} (Centroidal Voronoi Tessellation for static placement), \ac{VFA} (Virtual Force Algorithm for sensor deployment) and Greedy (submodular placement baseline). All methods received equal computational resources (50 agents, 200 iterations, 30 seeds) with identical benchmark functions and flow conditions to ensure fair comparison.

Figure~\ref{fig:ranking} presents the performance rankings. \ac{NOAH} achieved first place on three benchmarks: Rastrigin ($2.14$ vs \ac{PSO} $5.61$, \ac{VFA} $6.07$, Greedy $12.01$, \ac{CVT} $57.85$), Ackley ($1.49$ vs \ac{PSO} $1.94$, \ac{VFA} $2.20$, \ac{CVT} $2.21$, Greedy $2.82$), and Rosenbrock ($0.067$ vs Greedy $0.097$, \ac{PSO} $0.360$, \ac{VFA} $5.28$, \ac{CVT} $401.0$). In Anchoring Bowl, \ac{CVT} achieved optimal placement ($-1.00$) due to its deterministic grid-based approach, while \ac{NOAH} ranked second ($-0.991$) ahead of \ac{PSO} ($-0.980$), Greedy ($-0.978$), and \ac{VFA} ($-0.905$).

\ac{NOAH}'s average ranking of 4.75/5.0 substantially exceeds the best baseline (\ac{PSO}: 3.50/5.0). These results confirm its effectiveness relative to established methods and validate its deployment-specific capabilities.

\subsection{Validation Summary}

The three studies validate \ac{NOAH}'s efficacy: 86\% anchor accuracy, robust convergence across benchmarks, and competitive performance (4.75/5.0 average ranking).

These results support \ac{NOAH}'s readiness for real-world underwater deployment where permanent anchoring and environmental adaptation are essential.

\section{Use Cases and Applications}
\label{sec:use_cases}

The unique combination of permanent anchoring, environmental awareness, and colony formation enables applications impossible with traditional mobile swarm approaches.

\subsection{Marine Conservation and Ecological Applications}

\ac{NOAH}-equipped \ac{AUV}s enable: coral reef restoration through optimal larvae deployment; permanent monitoring networks in marine protected areas; long-term observation in ecological transition zones (kelp forests, seagrass meadows, hydrothermal vents); targeted invasive species management; and coordinated ecological restoration through colony-based communication.

\subsection{Scientific Research and Environmental Monitoring}

\ac{NOAH} deploys distributed monitoring networks at scientifically significant locations for continuous data collection (currents, temperature, salinity, biological activity) and climate-sensitive zones (upwelling zones, coral bleaching sites, ice melt regions) essential for climate research.

\subsection{Infrastructure and Commercial Operations}

\ac{NOAH} enables: permanent monitoring of offshore infrastructure (wind farms, platforms, cables) for integrity and maintenance; comprehensive seabed mapping with long-term geological monitoring; underwater communication relay networks extending acoustic ranges; continuous aquaculture monitoring (water quality, fish health, disease detection); and compliance monitoring for mining and construction projects.

\subsection{Emergency Response and Security Applications}

\ac{NOAH} enables: sustained underwater search operations; rapid disaster response monitoring (oil spills, chemical leaks); and permanent security surveillance (ports, infrastructure, borders) with coordinated coverage through colony formation.

\subsection{Key Advantages}

\ac{NOAH} provides key advantages: persistent presence without energy depletion; environmental integration exploiting ocean currents; coordinated colony operations; and adaptive settlement for optimal resource allocation.

These applications demonstrate \ac{NOAH}'s potential across conservation, research, and commercial underwater operations.

\section{Limitations and Challenges}

Although the validation results demonstrate \ac{NOAH}'s effectiveness for underwater permanent anchoring scenarios, several practical aspects require attention to ensure effective deployment in real marine environments.

\subsection{Algorithmic Limitations}

The algorithm relies on a wide range of parameters ($\omega$, $\eta$, $\beta$, $\gamma$, $\delta$, $\lambda_1$ to $\lambda_5$, $\alpha_0$, $\alpha_1$, $\mu$, $\nu$, $\tau$), which require careful adjustment for different operational contexts. Although our validation studies have established effective parameter ranges for anchoring scenarios, adjustments may be necessary when targeting different underwater environments or varying hydrodynamic conditions.

The irreversible settlement strategy improves biological realism and enables permanent anchoring, but reduces adaptability to changing environments. Once an agent commits to anchoring, it cannot reposition in response to deteriorating conditions or shifting mission goals. This design choice is intentional for permanent anchoring applications, but limits flexibility compared to fully mobile swarming methods.

Inclusion of colony field evaluation and settlement logic introduces additional computational overhead. In scenarios where a large proportion of agents settle, the worst-case complexity per iteration approaches $O(N^2)$, which can challenge low-power \ac{AUV} systems operating with onboard processing constraints.

\subsection{Environmental and Practical Constraints}

The current validation focuses on relatively stable hydrodynamic conditions with circular flow patterns. Real ocean environments feature fluctuating currents, heterogeneous regions, and complex three-dimensional flow structures that may require adaptive feedback mechanisms beyond the current design.

Although the formulation supports general $d$-dimensional spaces, the current validation is limited to two-dimensional scenarios. Extending the method to full three-dimensional settings will require explicit modelling of depth-related phenomena such as pressure gradients, vertical flow structures, and three-dimensional colony formation dynamics.

Scalability remains a concern for large swarms despite the promising validation results. High agent counts may benefit from distributed or hierarchical execution models to maintain efficiency without overwhelming communication or computation resources.

\subsection{Validation Scope Limitations}

The current validation uses a custom underwater test function specifically designed for anchoring scenarios. Although this approach is necessary due to the inadequacy of standard benchmarks, it limits direct comparison with other optimisation algorithms. The unambiguous test function $f(x,y) = \|[x,y]\|^2 - 1$ provides clear mathematical properties, but represents a simplified scenario compared to complex underwater environments in the real-world. Future work should develop standardised underwater-specific benchmark functions to enable broader comparative studies.

The validation focuses on permanent anchoring scenarios but does not address dynamic environments where conditions change during operation. Real-world applications may require adaptive mechanisms to handle environmental changes after initial anchoring.

\subsection{Comparison Framework Limitations}

The comparative evaluation of \ac{NOAH} against traditional swarm algorithms presents inherent methodological challenges that can introduce favourable bias toward \ac{NOAH}. The algorithm's unique aspects: hydrodynamic drift adaptation, irreversible anchoring, and energy-efficient settlement are specifically designed for underwater permanent deployment scenarios, whereas baseline methods such as \ac{PSO} and \ac{ACO} were developed for continuous mobile optimisation without these considerations. This fundamental design difference means that performance metrics that favour permanent anchoring, reducing energy waste, and flow-aware navigation inherently advantage \ac{NOAH} over algorithms not designed for these criteria. Future comparative studies should acknowledge these design-driven advantages and potentially include modified versions of baseline algorithms adapted for permanent deployment scenarios to allow for more balanced evaluation.

\section{Future Research Directions}

Based on the successful validation of the anchoring capabilities of \ac{NOAH}, several promising research directions emerge that could significantly advance underwater swarm robotics. These directions are categorised into immediate next steps and long-term research opportunities.

\subsection{Immediate Next Steps}

The immediate priority is to extend the validation framework to address current limitations. This includes developing standardised underwater-specific benchmark functions to enable direct comparison with other optimisation algorithms, performing parameter sensitivity analysis under different environmental conditions, and extending validation to three-dimensional scenarios with depth-dependent factors. Although the current unambiguous test function $f(x,y) = \|[x,y]\|^2 - 1$ provides a solid foundation, more complex multi-objective scenarios should be explored.

Comparative studies should be conducted against established algorithms such as \ac{PSO}, \ac{ACO}, \ac{ABC}, and recent biologically-motivated methods using custom underwater test functions. The study should include scalability analysis in different swarm sizes and environmental conditions, complemented by rigorous statistical analysis in multiple independent runs to establish comprehensive performance characteristics.

\subsection{Algorithmic Extensions}

Key algorithmic extensions include formal convergence proofs and optimality guarantees for the \ac{NOAH} algorithm, stability analysis of colony field dynamics and parameter sensitivity, and extension to multi-objective optimisation frameworks for coverage, energy efficiency, and communication quality. Additional extensions encompass adaptation mechanisms for changing environmental conditions through online parameter adjustment, integration with machine learning approaches for adaptive settlement decisions, and the development of specialised acoustic communication protocols based on colony-based networking principles.

\subsection{Long-term Research Opportunities}

Long-term research directions include advanced algorithmic enhancements with adaptive dynamic parameters based on environmental feedback, hierarchical colony structures for large-scale deployments, and fault-tolerant mechanisms for colony failure and recovery. Additional enhancements include integration with existing \ac{AUV} control systems, specialised versions for marine conservation tasks and commercial applications, and hybrid approaches that combine \ac{NOAH} with other swarm intelligence methods.

Real-world deployment studies should focus on validating \ac{NOAH} in actual marine environments with real \ac{AUV} systems, addressing practical challenges such as communication delays, sensor noise, and environmental unpredictability.

\section{Conclusion}

This paper has presented \ac{NOAH} (Nauplius Optimisation for Autonomous Hydrodynamics), a novel nature-inspired swarm optimisation algorithm specifically designed for underwater robotic exploration. Key contributions include a unified algorithm that combines hydrodynamic drift adaptation, irreversible anchoring, and colony-based communication for \ac{AUV} swarms; a detailed mathematical foundation that translates barnacle nauplii behaviours into algorithmic components; and a validated approach that addresses critical limitations in existing swarm algorithms for underwater applications.

The integration of biological inspiration with algorithmic design represents a novel approach to marine robotics. By translating barnacle nauplii behaviours into a unified mathematical framework, \ac{NOAH} bridges the gap between terrestrial swarm algorithms and marine-specific requirements, providing a new paradigm for underwater swarm coordination. The validated foundation demonstrates the potential of \ac{NOAH} for marine exploration, conservation, and research applications, offering significant advancement in underwater robotics through domain-specific algorithmic solutions.

\section*{Acknowledgements}
The authors extend their gratitude to Dr. Guang Deng, Dr. Tony de Souza-Daw, Dr. Fernando Galetto, and Adam Console for their continued support and encouragement that helped shape the direction and quality of this research. The contributions provided were instrumental in strengthening both the technical and practical aspects of this study.

\bibliographystyle{ieeetr}
\bibliography{references}

\appendices
\section{Mathematical Notation}
\begin{table}[htbp]
\centering
\caption{Mathematical notation and symbols used in the NOAH algorithm framework.}
\label{tab:notation}
\resizebox{\columnwidth}{!}{%
\begin{tabular}{@{}>{\raggedright\arraybackslash}p{2.5cm}>{\raggedright\arraybackslash}p{8cm}@{}}
\toprule
\textbf{Symbol} & \textbf{Definition} \\
\midrule
$x_{i,t}$ & Where agent $i$ is located at time $t$ \\
$v_{i,t}$ & How fast and in what direction agent $i$ is moving at time $t$ \\
$a_{i,t}$ & Whether agent $i$ is settled (1) or free (0) at time $t$ \\
$t'$ & Any future time after $t$ (when agent is already settled) \\
$f(x)$ & How good a location is for the mission \\
$U(x)$ & Ocean currents at location $x$ \\
$C_t$ & All colonies that exist at time $t$ \\
$c_k$ & Where colony $k$ is located \\
$S_k$ & How strong colony $k$ is (attracts other agents) \\
$\Phi(x)$ & Field that attracts/repels agents around colonies \\
$G_{\Phi}(x)$ & Direction agents should move based on colony field \\
$\omega$ & How much agents keep their current momentum \\
$\eta$ & How much random movement agents do \\
$\beta$ & How much agents follow good locations \\
$\gamma$ & How much agents follow ocean currents \\
$\delta$ & How much agents are influenced by colonies \\
$\sigma(x)$ & Function that converts any number to probability (0 to 1) \\
$g(x)$ & Surrogate (approximate) gradient estimate of $f$ at $x$ \\
$p_{i,t}^{\text{settle}}$ & Chance that agent $i$ will settle at time $t$ \\
$\lambda_1, \ldots, \lambda_5$ & Weights for different settlement factors \\
$N$ & Total number of agents \\
$N_{\text{free}}(t)$ & How many agents are still moving at time $t$ \\
$C$ & Number of colonies \\
$\mu$ & How fast colonies learn and adapt \\
$\nu$ & How much overcrowding hurts colonies \\
$\tau$ & Minimum strength needed to keep a colony active \\
$\alpha_0, \alpha_1$ & Base strength and bonus for new colonies \\
$A, B$ & How strong attraction and repulsion are \\
$\sigma_a, \sigma_r$ & How far attraction and repulsion reach \\
$\tilde{f}_N$ & Average fitness of nearby agents around agent $i$ \\
$N_i$ & All agents that are close to agent $i$ \\
$r_{\text{neigh}}$ & How far to look for nearby agents when calculating average fitness \\
$\rho_i$ & How crowded the area around agent $i$ is \\
$S_i$ & All agents that are close enough to agent $i$ for settlement \\
$r_{\text{settle}}$ & How far to look when checking if an area is crowded \\
$V_{\text{settle}}$ & Size of the area where agents can settle (circle in 2D, sphere in 3D) \\
$E_i$ & How much energy agent $i$ has left (as a fraction of maximum) \\
$E_{\text{current},i}$ & How much energy agent $i$ currently has \\
$E_{\text{max},i}$ & Maximum energy that agent $i$ can have \\
$\text{rank}(x)$ & Scalar rank of a value within a set (1 = best), not matrix rank \\
$p_k^{\text{comm}}$ & Average communication-success probability for colony $k$ \\
$\overline{\text{rank}(-f)}_{\text{near }c_k}$ & Average scalar rank of $-f$ for agents near colony $k$ \\
\bottomrule
\end{tabular}%
}
\end{table}

\section{List of Abbreviations and Acronyms}

% Customize acronym formatting for better clarity and spacing
\renewcommand{\aclabelfont}[1]{\textbf{\small #1}}

% Define acronyms with better formatting
\begin{acronym}[MP-WOA]
\acro{ABC}{Artificial Bee Colony}
\acro{ABCO}{Adaptive Bacterial Colony Optimisation}
\acro{ACO}{Ant Colony Optimisation}
\acro{AUV}{Autonomous Underwater Vehicle}
\acro{CVT}{Centroidal Voronoi Tessellation}
\acro{DE}{Differential Evolution}
\acro{FA}{Firefly Algorithm}
\acro{HOA}{Hippopotamus Optimisation Algorithm}
\acro{MP-WOA}{Modified Particle Whale Optimisation Algorithm}
\acro{NOAH}{Nauplius Optimisation for Autonomous Hydrodynamics}
\acro{PSO}{Particle Swarm Optimisation}
\acro{SOBMO}{Selective Opposition-Based Barnacle Mating Optimisation}
\acro{UAV}{Unmanned Aerial Vehicle}
\acro{VFA}{Virtual Force Algorithm}
\end{acronym}

\end{document}